# Transparent, Evaluable, and Accessible Data Agents: A Proof-of-Concept Framework

Nooshin Bahador

University Health Network (UHN), Toronto, ON, Canada

**ABSTRACT**

This article presents a modular, component-based architecture for developing and evaluating AI agents that bridge the gap between natural language interfaces and complex enterprise data warehouses. The system directly addresses core challenges in data accessibility by enabling non-technical users to interact with complex data warehouses through a conversational interface, translating ambiguous user intent into precise, executable database queries to overcome semantic gaps. A cornerstone of the design is its commitment to transparent decision-making, achieved through a multi-layered reasoning framework that explains the "why" behind every decision, allowing for full interpretability by tracing conclusions through specific, activated business rules and data points. The architecture integrates a robust quality assurance mechanism via an automated evaluation framework that serves multiple functions: it enables performance benchmarking by objectively measuring agent performance against golden standards, and it ensures system reliability by automating the detection of performance regressions during updates. The agent's analytical depth is enhanced by a statistical context module, which quantifies deviations from normative behavior, ensuring all conclusions are supported by quantitative evidence including concrete data, percentages, and statistical comparisons. We demonstrate the efficacy of this integrated agent-development-with-evaluation framework through a case study on an insurance claims processing system. The agent, built on a modular architecture, leverages the BigQuery ecosystem to perform secure data retrieval, apply domain-specific business rules, and generate human-auditable justifications. The results confirm that this approach creates a robust, evaluable, and trustworthy system for deploying LLM-powered agents in data-sensitive, high-stakes domains.

## OPEN-SOURCE IMPLEMENTATION

All code related to this work is available in the accompanying **GitHub repository**.

## 1. INTRODUCTION

The "black box" nature of artificial intelligence (AI) remains one of the challenges hindering its widespread adoption, particularly in high-stakes domains such as healthcare, finance, and legal decision-making. This opacity in AI systems, wherein the internal workings and rationale behind decisions are not transparent, raises concerns about trust, accountability, and ethical use. Numerous studies emphasize that for AI to be reliably integrated and accepted, it is crucial that its decision-making processes become interpretable and accessible to users (Nunes et al., 2024).

Addressing this need for transparency and interpretability, recent advancements have introduced a general framework for large language model (LLM)-based agents, composed of three core components: Brain, Perception, and Action. The Brain, powered by LLMs, handles reasoning, planning, knowledge retrieval, and natural language interaction, making these models well-suited to serve as the cognitive core of intelligent agents. The Perception component processes multimodal inputs, enabling the agent to interpret diverse data sources beyond text. Meanwhile, the Action component focuses on executing tasks, often by leveraging external tools such as calculators, APIs, and databases to address inherent limitations of LLMs, such as knowledge cutoffs and gaps in specialized expertise. Furthermore, effective agents maintain memory by storing past interactions, thoughts, and actions, which can be summarized, compressed, or retrieved to inform future decision-making and improve performance over time (Xi et al., 2023).

While these frameworks show promise, recent research highlights an urgent need to evaluate large language models (LLMs) as agents operating within complex, interactive environments. Such evaluations help reveal their true capabilities and limitations. A key gap lies in the shift from subjective evaluations to more quantitative assessments of agent behavior, especially in domains like database querying where failures such as "Invalid Format" and "Invalid Action" frequently occur. These errors often stem from weak reasoning and poor adherence to task constraints. Despite improvements, LLM agents continue to struggle with long-term reasoning, consistent decision-making, and following multi-step instructions, critical aspects for real-world deployment. As such, achieving high-quality alignment remains a central challenge and a necessary step toward creating more capable and reliable LLM-based agents (Liu et al., 2023).

In parallel with agent evaluation, recent research suggests that a strong large language model (LLM) can serve as a scalable and accurate proxy for human evaluation in chatbot assessments. Traditional benchmarks often fall short in capturing the qualities that make a chatbot truly helpful or aligned with human preferences. Instead, more meaningful measures of success include the model's ability to follow instructions and perform well in multi-turn conversations (Zheng et al., 2023).

To assess these conversational capabilities, researchers design sets of multi-turn questions that test both instruction-following and conversational quality. However, studies have shown that LLM-based evaluations can introduce certain biases. These include: 1) Position bias, or a preference for the first answer presented; 2) Verbosity bias, favoring longer or more detailed responses even when unnecessary; and 3) Limited reasoning, reflecting challenges with mathematical or logical reasoning tasks. Understanding and mitigating these biases is critical to ensuring fair and accurate evaluation of conversational AI systems (Zheng et al., 2023).

To further enhance reasoning and interpretability within LLMs, chain-of-thought (CoT) prompting has emerged as a significant advancement. This technique offers a structured approach to reasoning by encouraging the model to generate intermediate steps prior to arriving at a final answer. Chain-of-thought prompting has been shown to improve performance across multiple reasoning-intensive tasks and serves as an interpretable mechanism to better understand the internal reasoning processes of LLMs. Research demonstrates CoT's effectiveness in three key domains: arithmetic reasoning (e.g., solving math word problems), commonsense reasoning (e.g., answering questions requiring implicit world knowledge), and symbolic reasoning (e.g., tasks like last letter concatenation). By breaking down complex problems into step-by-step thought processes, CoT prompting not only enhances model accuracy but also provides insights into the model's strengths and limitations, positioning it as a foundational technique for improving both transparency and performance in LLM-based systems (Wei et al., 2022).

This work directly addresses these critical gaps by implementing a modular, LLM-based agent architecture that operationalizes transparency and enables quantitative evaluation. By designing a system with a multi-layered interpretability framework, we force the agent's "Brain" to explicitly articulate its reasoning process (breaking down decisions into input analysis, rule activation, and decision pathways) thereby demystifying the "black box" and making the rationale behind high-stakes decisions fully auditable. Furthermore, we integrated automated evaluation framework directly confronts the need for objective assessment of agent behavior; it systematically benchmarks performance against golden standards, measures specific failure modes like reasoning consistency and instruction-following.

## 2. METHOD

Designing an effective agentic system requires an approach that integrates several core components working in harmony. At the foundation lies the agent definition and configuration, where the system's purpose is established through clear instructions, integrate essential tools like BigQuery data sources and external APIs, and select the appropriate LLM model such as Gemini that aligns with the specific use case requirements. This agent core then connects to robust data infrastructure involving properly structured BigQuery datasets with secure credential management, efficient query execution capabilities, and metadata management to ensure the system understands schema relationships and data context.

The system architecture follows a logical flow where user questions trigger the agent to generate SQL queries, execute them against BigQuery, and formulate natural language responses, while simultaneously feeding into an evaluation framework that compares outputs against golden answers from test datasets. This evaluation component is critical for maintaining quality assurance, enabling performance benchmarking between agent versions, and detecting regressions during updates through automated testing pipelines. Key design decisions involve customizing domain-specific elements like agent instructions, tools, and evaluation datasets while reusing the underlying Agent Development Kit (ADK) framework, evaluation infrastructure, and project scaffolding to maintain consistency and efficiency.

The implementation process follows a structured workflow that begins with infrastructure setup, where BigQuery dataset is prepared, necessary Google Cloud APIs are enabled, and authentication protocols are configured. The workflow then progresses to agent development, during which system instructions are defined, tools are configured, and initial testing phases are conducted. The process continues with the creation of evaluation datasets containing real user questions and golden answers that cover various edge cases. This is followed by running automated evaluations to identify weaknesses and drive iterative improvements throughout the development lifecycle. Success in this approach hinges on multiple critical factors, including maintaining high data quality through clean, well-documented datasets; implementing thoughtful agent design with clear instructions and robust error handling mechanisms; and applying evaluation practices using diverse test cases with actionable metrics. This entire methodology operates within an iterative development cycle encompassing design, implementation, evaluation, and improvement phases.

Figure 1 presents a structured flowchart of an AI-powered insurance claims processing system, integrating data validation, decision-making, and quality assurance. The workflow begins with claims retrieved from a database, which are then validated before entering a multi-component reasoning engine. This engine comprises modules such as the Business Rule Engine, Pattern Detection Module, and Statistical

Comparator, each referencing a centralized business rules repository. Based on confidence scoring, the system determines whether a claim is an outlier and requires human review or can be auto-approved. Additionally, monitoring, evaluation, and conversation management systems feed into quality control loops to ensure performance improvement.

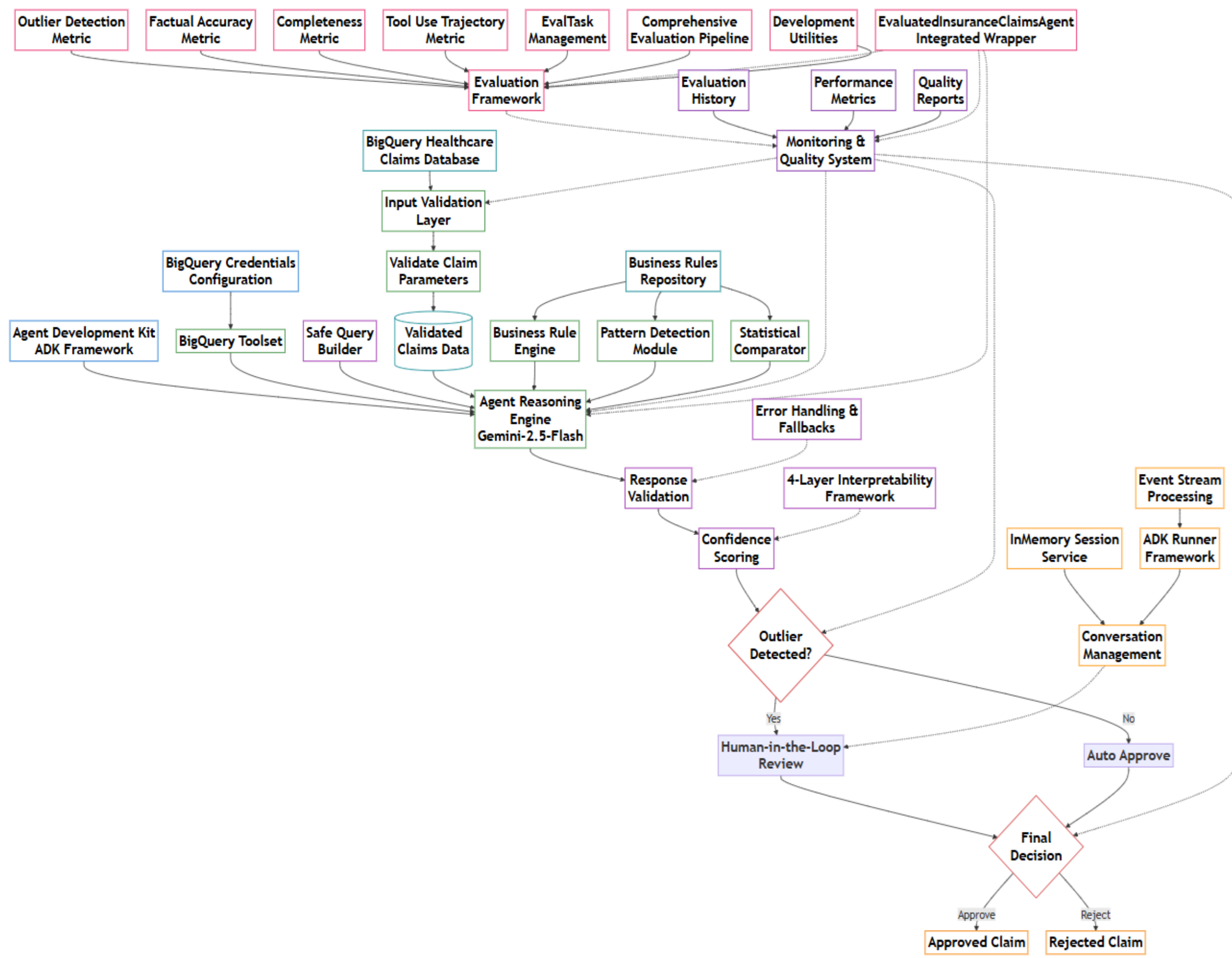

**Figure 1.** This figure presents a system architecture diagram for the AI agent, illustrating the complete end-to-end pipeline from data input through final decision-making. The system begins with multiple evaluation metrics at the top (Outlier Detection, Factual Accuracy, Completeness, Tool Use Trajectory, etc.) that feed into an Evaluation Framework connected to BigQuery Database, which serves as the central data repository. The core processing flow shows data moving through an Input Validation Layer that validates claim parameters, followed by a Business Rules Repository that applies domain-specific logic for detecting unusual diagnosis-procedure combinations, abnormally high claim amounts, and geographic restrictions. The validated claims data then passes through an Agent Reasoning Engine powered by Gemini-2.5-Flash, which incorporates multiple analytical components including a Business Rule Engine, Pattern Detection Module, and Statistical Comparator to perform multi-layered reasoning analysis. The system includes robust quality assurance through Response Validation, Confidence Scoring, and a 4-Layer Interpretability Framework, while the Agent Development Kit (ADK Framework) provides the underlying infrastructure with BigQuery Toolset, Safe Query Builder, and other development utilities.

The dataset represents synthetic health insurance claims data generated for telehealth services in the year 2024, including fields such as claim ID, patient ID, provider name, procedure type, diagnosis, claim amount, claim date, and status indicators. The data simulates scenarios across various procedure types like Virtual Consultation, Mental Health Session, Prescription Refill, Follow-up Visit, and Emergency Consult, each with an associated average cost and duration. Outlier detection is built into the data generation logic to identify potentially suspicious claims based on four key rules: unusual diagnosis-procedure combinations (e.g., Mental Health Session with Common Cold), excessively high costs (more than three times the average for a given procedure), geographic restrictions (telehealth services in unsupported states), and randomly introduced suspicious patterns to mimic fraud. Any claim identified as an outlier is flagged for review, and its status is set to "Pending"; otherwise, it is automatically marked as "Approved". The **Table 1** describes each field in the insurance claims dataset.

**Table 1. Insurance Claims Dataset - Data Dictionary**

| Field Name | Data Type | Description | Example Value | Notes |
|---|---|---|---|---|
| claim_id | String | Unique identifier for each insurance claim. | CLM_10000 | Always starts with "CLM_" followed by a unique number. |
| patient_id | String | Unique identifier for the patient. | PAT_2824 | Format: PAT_XXXX |
| provider_name | String | Name of the healthcare provider. | TeleHealth Inc | One of: TeleHealth Inc, VirtualCare Co, RemoteMed Group, DigitalDoc Services |
| procedure_type | String | Type of healthcare procedure performed. | Prescription Refill | E.g., Virtual Consultation, Emergency Consult, Follow-up Visit |
| diagnosis | String | Diagnosis associated with the claim. | Depression | E.g., Back Pain, Anxiety, Migraine, etc. |
| claim_amount | Float | Dollar amount being claimed. | 57.45 | Varies based on procedure and cost variation. |
| claim_date | Date (String) | Date when the claim was made. | 2024-03-12 | Random date within 2024. |
| patient_state | String | U.S. state code where the patient resides. | FL | E.g., CA, NY, TX, FL, etc. |
| is_outlier | Boolean | Indicates if the claim is considered an outlier based on certain rules. | false | Determined based on cost, diagnosis-procedure mismatch, location, etc. |
| outlier_reason | String (nullable) | Explanation for outlier classification (if applicable). | null | Values like "Abnormally high claim amount", or null if not an outlier. |
| review_required | Boolean | Flag indicating whether manual review is required. | false | Matches is_outlier. |
| claim_status | String | Current status of the claim. | Approved | Either Approved or Pending (if outlier). |

The insurance claims agent is architected as a specialized, rule-driven analytical system designed to evaluate healthcare claims related to virtual visits by interfacing directly with a secured BigQuery healthcare claims database. It integrates robust input validation mechanisms to ensure data integrity and applies a set of domain-specific business rules to detect outliers based on unusual diagnosis-procedure combinations, abnormal claim amounts, geographic inconsistencies, and suspicious claim patterns. The agent decomposes the decision process into input analysis, rule activation, decision pathways, and confidence scoring, thereby enabling transparent, explainable flagging of claims for human review. This modular design emphasizes patient privacy, restricts scope to predefined datasets, and supports iterative pattern detection, root cause analysis, and rule effectiveness evaluation. **Table 2** describes the internal architecture and reasoning framework of the insurance claims agent.

**Table 2. Functional Design of the Claims Evaluation Agent**

| Component | Purpose | Design Logic / Reasoning |
|---|---|---|
| Agent Identity | Embeds a clearly defined role and scope | Operates as a specialized analytical agent for virtual healthcare claims; constrained to the dataset to enforce domain containment and reduce unintended behavior. |
| Data Access Interface | Secure and controlled retrieval of relevant claim records | Employs parameterized, role-limited querying to access only the insurance claims table; ensures query safety and data minimization. |
| Input Validation Layer | Early-stage filtering of malformed or suspicious data | Enforces domain bounds (e.g., cost range, date recency, valid state codes, known procedures) to prevent garbage-in-garbage-out errors. |
| Business Rule Engine | Encodes domain-specific outlier detection logic | Applies interpretable rules involving cost thresholds, unusual diagnosis-procedure pairings, and geo-restrictions to systematically flag anomalies. |
| Multi-layered Reasoning | Transparent explanation of agent decisions | Structured across four reasoning layers: (1) Input Feature Analysis, (2) Rule Activation, (3) Decision Pathway, and (4) Confidence Scoring for traceable logic. |
| Interpretable Output | Generates human-auditable justifications | Produces structured explanations including rule violations, statistical deviations, and recommendations to support downstream human review. |
| Pattern Detection Module | Detects behavioral anomalies across claims | Identifies systemic patterns such as repetitive high-value claims, non-business hour submissions, and regional fraud indicators. |
| Statistical Comparator | Quantifies deviation from normative behavior | Compares individual claims against aggregated historical distributions to assess likelihood of abnormality. |
| Confidence Scoring | Quantifies trust in anomaly detection | Uses heuristics and rule interactions to express uncertainty (High/Medium/Low) and signal when human intervention is essential. |
| Human-in-the-Loop Design | Ensures ethical and accountable decision-making | Avoids final decisions on claim validity; flags suspicious entries and explicitly recommends review without making autonomous determinations. |
| Audit Support Layer | Facilitates transparency and traceability for compliance | Provides templated, reproducible justifications suitable for audits, appeals, or regulatory reporting. |
| Toolset Integration | Connects to external data processing tools | Interfaces with secure analytical backends (e.g., cloud data warehouses) to enable real-time validation, comparison, and decision support. |

**Table 3** abstractly represents the backend design of the asynchronous conversational interface responsible for managing interactions with the agent referenced in Table 2. This architecture handles the full conversation lifecycle: initiating sessions, sending user prompts, processing agent-generated streamed events (including function/tool calls), and validating final outputs before returning a structured result.

**Table 3. Asynchronous Conversation Lifecycle Management for Insurance Claims Agent**

| Component/Concept | Description | Role in Conversation Lifecycle |
|---|---|---|
| Session Management | Manages unique conversation sessions in memory, identified by session and user IDs | Creates and tracks the lifecycle of a conversation session |
| Agent Initialization | Loads or creates the specialized domain agent (e.g., insurance claims agent) | Provides the conversational AI logic tailored to the use case |
| Runner | Orchestrates the asynchronous interaction between user input, agent processing, and output | Runs the agent asynchronously, handling streaming events |
| User Prompt Handling | Accepts user input to be sent as a message to the agent | Initiates conversation or sends new messages during a session |
| Event Streaming | Processes streamed events from the agent, which may include partial responses or function calls | Allows real-time, incremental response handling from the agent |
| Function Call Detection | Detects and extracts agent-initiated function/tool calls embedded in streamed events | Captures actions or external tool invocations triggered by the agent |
| Final Response Extraction | Gathers the final textual response after agent finishes processing | Provides the conclusive answer or output of the conversation |
| Response Validation | Checks final agent response for required content sections and sensitive information (e.g., PII) | Ensures response integrity and compliance before presenting to user |
| Error Handling | Catches exceptions during conversation flow | Prevents crashes and provides fallback error messages |
| Output Structure | Returns a structured result including response text, detected tool calls, and validation status | Supplies data about the conversation outcome |

**Table 4** presents an abstracted view of the evaluation framework designed to assess the performance of a conversational AI agent in structured experiments. The framework automates the end-to-end evaluation process by invoking the agent with prompts, collecting responses, applying a scoring system, and saving structured metrics. It supports both summary-level insights (e.g., average accuracy and completeness) and detailed, per-instance pointwise analysis. Results are serialized into reproducible output files to enable longitudinal tracking of model performance across different experimental runs. The design ensures transparency, repeatability, and interpretability of agent behavior through quantitative benchmarks.

**Table 5** outlines the structure of the evaluation dataset used to benchmark the performance of the insurance claims conversational agent. The dataset consists of designed prompt-reference pairs, where each prompt represents a query that a domain analyst might ask, and each reference provides the expected content or reasoning that the agent should ideally return. These pairs serve as ground truth for evaluating the agent's factual accuracy, completeness, and interpretability. The dataset is diverse in scope, covering statistical queries, pattern detection, rule-based explanations, and exception handling, ensuring coverage of the agent's capabilities across functional dimensions.

**Table 4. Evaluation Framework for Conversational Agent Performance**

| **Component** | **Purpose** | **Design Role in Evaluation Lifecycle** |
|---|---|---|
| Agent Invocation Layer | Sends prompt to the conversational agent and retrieves structured response | Initiates evaluation by simulating real user interactions |
| Error Handling | Catches and reports failures in agent execution | Ensures robust execution without halting the entire evaluation process |
| Result Serialization | Converts evaluation results into a consistent JSON format | Enables reproducibility and persistent storage of experiment outputs |
| Directory Management | Organizes results by experimental run identifier | Maintains clean separation and traceability of individual evaluation runs |
| Summary Metric Reporter | Prints high-level metrics such as average accuracy and completeness | Provides quick insight into overall agent quality |
| Pointwise Metrics Processor | Aggregates per-example scores for granular analysis | Supports diagnostic evaluations across a diverse set of queries |
| Data Aggregation Layer | Computes averages and aggregates for metric reporting | Quantifies general trends across multiple test cases |
| Output Presentation | Formats and displays evaluation results in human-readable form | Improves interpretability and supports analyst decision-making |

**Table 5. Structure of the Evaluation Dataset for Insurance Claims Agent**

| Dataset Element | Purpose | Example Scenario |
|---|---|---|
| Prompt | Represents a user's natural language query to the agent | "Which providers have the highest outlier rates?" |
| Reference Response | Describes the expected or ideal content that the agent should return | "The analysis should show provider names and their respective outlier percentages." |
| Outlier Reasoning | Tests the agent's ability to explain anomalies based on encoded business rules | "Explain why claim CLM_10050 was flagged as an outlier." |
| Statistical Aggregation | Evaluates the agent's handling of numeric queries and averages | "What is the average claim amount for Virtual Consultation procedures that are NOT outliers?" |
| Categorical Breakdown | Assesses the agent's ability to segment data by categories or dimensions | "Which states have the highest rate of geographic mismatch outliers?" |
| Rule-based Explanation | Validates interpretability and transparency of decision logic | "What is the mechanistic interpretability behind flagging claim CLM_10100?" |

| Dataset Element | Purpose | Example Scenario |
|---|---|---|
| Quantitative Summary | Checks for correct summarization of counts or percentages | “How many claims require human review currently?” |
| Comparative Distribution | Challenges the agent to compare value ranges across groups | “Show the distribution of claim amounts for outlier vs non-outlier claims.” |
| Behavioral Anomaly Detection | Targets detection of mismatched or suspicious patterns | “Show me claims that were flagged for unusual diagnosis-procedure combinations.” |
| Coverage Completeness | Measures breadth and accuracy of the agent’s response scope | “What are the most common outlier reasons in the insurance claims dataset?” |

**Table 6** illustrates the evaluation metrics framework designed to assess the performance of a conversational agent specialized in insurance claims analysis. The framework uses structured pointwise metrics to rate the agent's responses based on critical dimensions such as outlier detection accuracy, factual correctness, and response completeness. Each metric is grounded in a set of criteria and a standardized rating rubric to ensure consistency and objectivity. The evaluation process runs over a curated dataset of prompt-reference pairs, producing quantitative scores that reflect how well the agent adheres to business rules, explains its reasoning, and delivers informative, accurate answers.

**Table 6. Evaluation Metrics Framework for Insurance Claims Conversational Agent**

| **Metric Name** | **Purpose** | **Evaluation Criteria** | **Rating Rubric (Example)** |
|---|---|---|---|
| Outlier Detection Metric | Measures correctness and clarity in identifying claim outliers | - Detection Accuracy<br>- Interpretability of explanation | 5 = Perfect detection with clear reasoning<br>3 = Mostly correct<br>1 = Incorrect or unclear |
| Factual Accuracy Metric | Assesses alignment between agent response and known reference facts | - Accuracy of information | 5 = All facts match<br>3 = Minor inaccuracies<br>1 = Major factual errors |
| Completeness Metric | Evaluates whether the response includes all requested or relevant details | - Inclusion of all key information | 5 = Fully complete<br>3 = Minor omissions<br>1 = Critically incomplete |
| Metric Prompt Template | Provides standardized evaluator instructions and input variables | - Prompt, reference, and response used as evaluation context | Ensures consistent scoring across examples |
| Evaluation Task Runner | Executes the evaluation pipeline over a dataset using selected metrics | - Applies multiple metrics across all prompt-response pairs | Enables reproducible, structured performance evaluation |
| Scoring Aggregator | Summarizes and stores evaluation results | - Produces summary and pointwise metrics saved per experiment | Supports auditability and longitudinal performance tracking |
| Experiment Identifier | Tracks evaluation run using timestamps and UUIDs | - Unique ID per run | Facilitates versioning and experiment reproducibility |

## 3. RESULTS

**Figure 2** presents the BigQuery database schema interface for the healthcare claims dataset that underpins the AI agent system. The left panel shows the project structure, with the dataset insurance_claims and its claims_data table located under the project ID woven-invention-469721-s1, which serves as the primary data source for the agent's queries. The right panel displays the full schema of the claims_data table, consisting of 12 fields including claim_status, review_required, is_outlier (all BOOLEAN), claim_date (DATE), claim_amount (FLOAT), outlier_reason (STRING), and several other STRING fields such as diagnosis, provider_name, claim_id, patient_state, patient_id, and procedure_type. The interface indicates an active BigQuery environment with capabilities to query and explore the dataset, reflecting the live infrastructure used by the agent for real-time SQL execution.

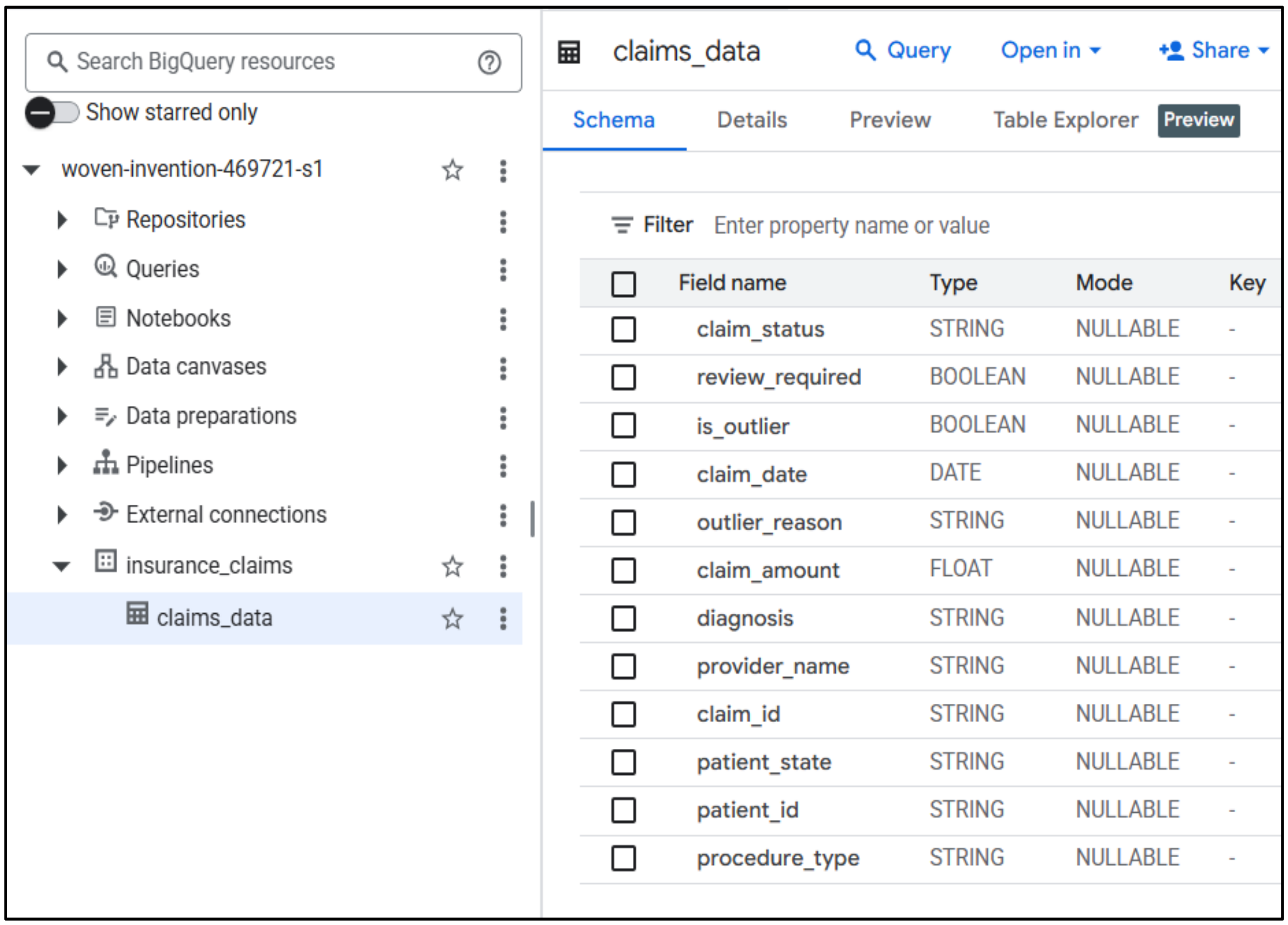

**Figure 2.** BigQuery schema interface showing the 12-field structure of the healthcare claims database that powers the AI agent's capabilities.

**Figure 3** illustrates a two-part interaction between a user and an agent, demonstrating how natural language queries are accurately translated into real-time database lookups. On the left, the user asks, “What was a sample claim amount for stomach flu in New York?” and the agent responds with, “A sample claim amount for stomach flu in New York was $330.76”, based on a SQL query filtering for diagnosis = 'Stomach Flu' and patient_state = 'NY'. When the user follows up with, “Give me the claim ID for that”, the agent replies, “The claim ID for that sample is CLM_10386”. On the right, the underlying database records accessed by the agent are shown, confirming the accuracy of the response: both entries for CLM_10386 correspond to an Emergency Consult for Stomach Flu in NY with a claim amount of $330.76. These records match the agent’s response exactly, validating the correctness of the executed query. This alignment between the agent's output and the database confirms the reliability of the conversational interface in retrieving precise, traceable information.

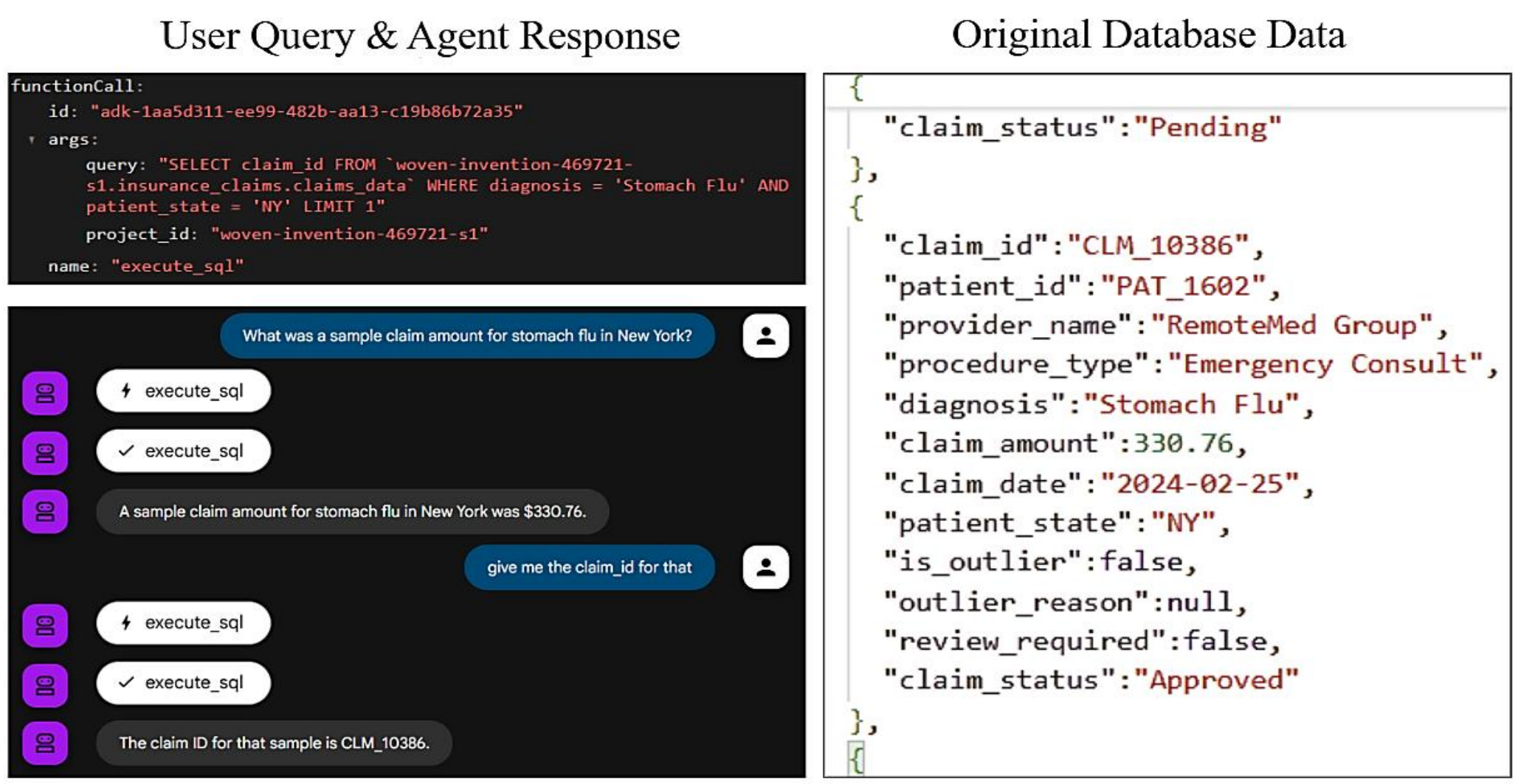

**Figure 3.** Agent-database alignment demonstration

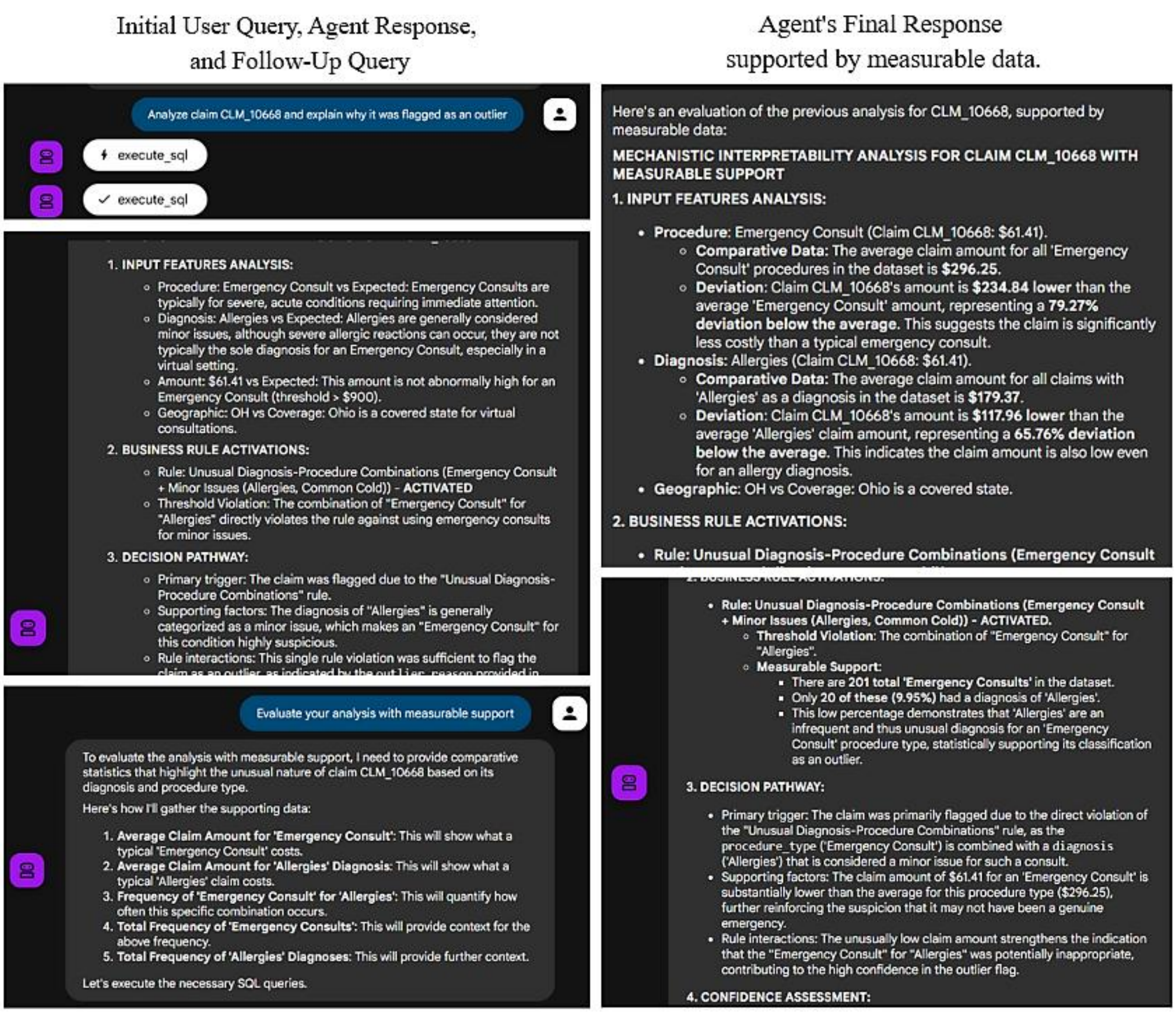

**Figure 4.** The agent demonstrates statistical reasoning by comparing its response against dataset-wide distributions, providing quantitative evidence to support its conclusions.

**Figure 4** demonstrates the agent's reasoning capabilities through a detailed analysis of claim CLM_10668, showcasing how it combines statistical analysis with business rule interpretation to provide explanations for outlier detection. The left panel shows the initial user query asking to analyze why CLM_10668 was flagged as an outlier, followed by the agent's systematic approach involving input features analysis (Emergency Consult procedure vs expected norms), business rule activations (specifically the "Unusual Diagnosis-Procedure Combinations" rule for Emergency Consult + Allergies), and decision pathway reasoning that explains how this single rule violation was sufficient to flag the claim. The right panel reveals the agent's comprehensive final response supported by quantitative evidence: it calculates that CLM_10668's amount of $61.41 represents a 67.47% deviation below the average Emergency Consult cost of $296.25, compares this against the dataset's statistical distributions (showing only 20 of 201 total Emergency Consults had an 'Allergies' diagnosis, representing just 9.95%), and provides confidence assessment based on measurable support rather than subjective judgment. The agent's reasoning integrates multiple data points (procedure type averages, diagnosis frequency statistics, rule threshold violations, and comparative analysis) to deliver an interpretable explanation that moves beyond simple rule-matching to demonstrate analytical reasoning with numerical precision, ultimately concluding that while the claim amount itself wasn't suspicious, the unusual pairing of Emergency Consult with Allergies diagnosis triggered the outlier classification due to its statistical rarity in the dataset.

**Figure 5** shows the distributed microservices architecture starting up to support the conversational AI interface, with each service handling different aspects. Each line shows HTTP server startup information with timestamps, IP addresses (127.0.0.1), different port numbers (64872, 64896, etc.), and GET request paths that include session IDs and various endpoints for agent interactions, trace collection, and graph generation.

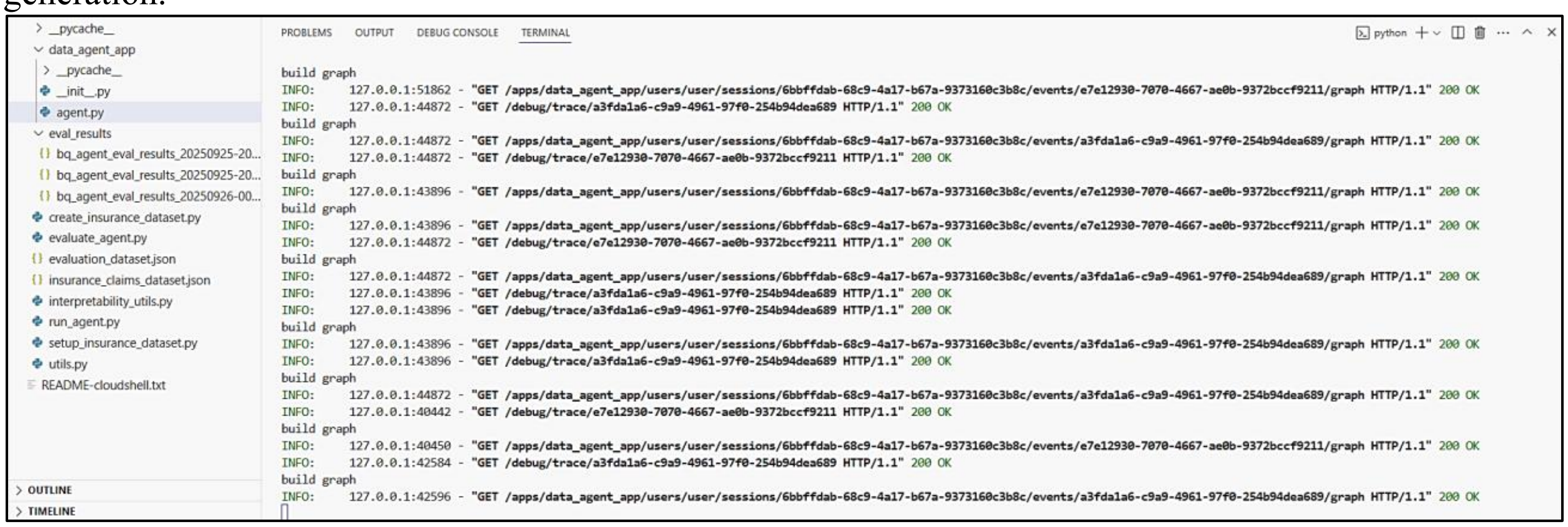

**Figure 5.** Distributed microservices architecture startup log for a healthcare claims AI system, showing successful initialization of agent services, evaluation frameworks, and data processing components across multiple ports.

**Figure 6** illustrates how multiple interconnected components work together to deliver intelligent claims processing. The top panel displays the Agent Development Kit interface, where a user submits a query and the agent responds. The middle section reveals three identical function call graphs that show the systematic execution pattern: the insurance_claims_agent makes sequential API calls to gather metadata before executing the query. The bottom panels expose the technical infrastructure powering this interaction. The functionCall section shows the SQL query filtering process, while the functionResponse section presents the successful execution results. This system seamlessly integrates a conversational AI interface with robust backend data processing. Natural language queries are automatically translated into secure database operations, business rules are systematically applied, and structured results are returned with full transparency and auditability.

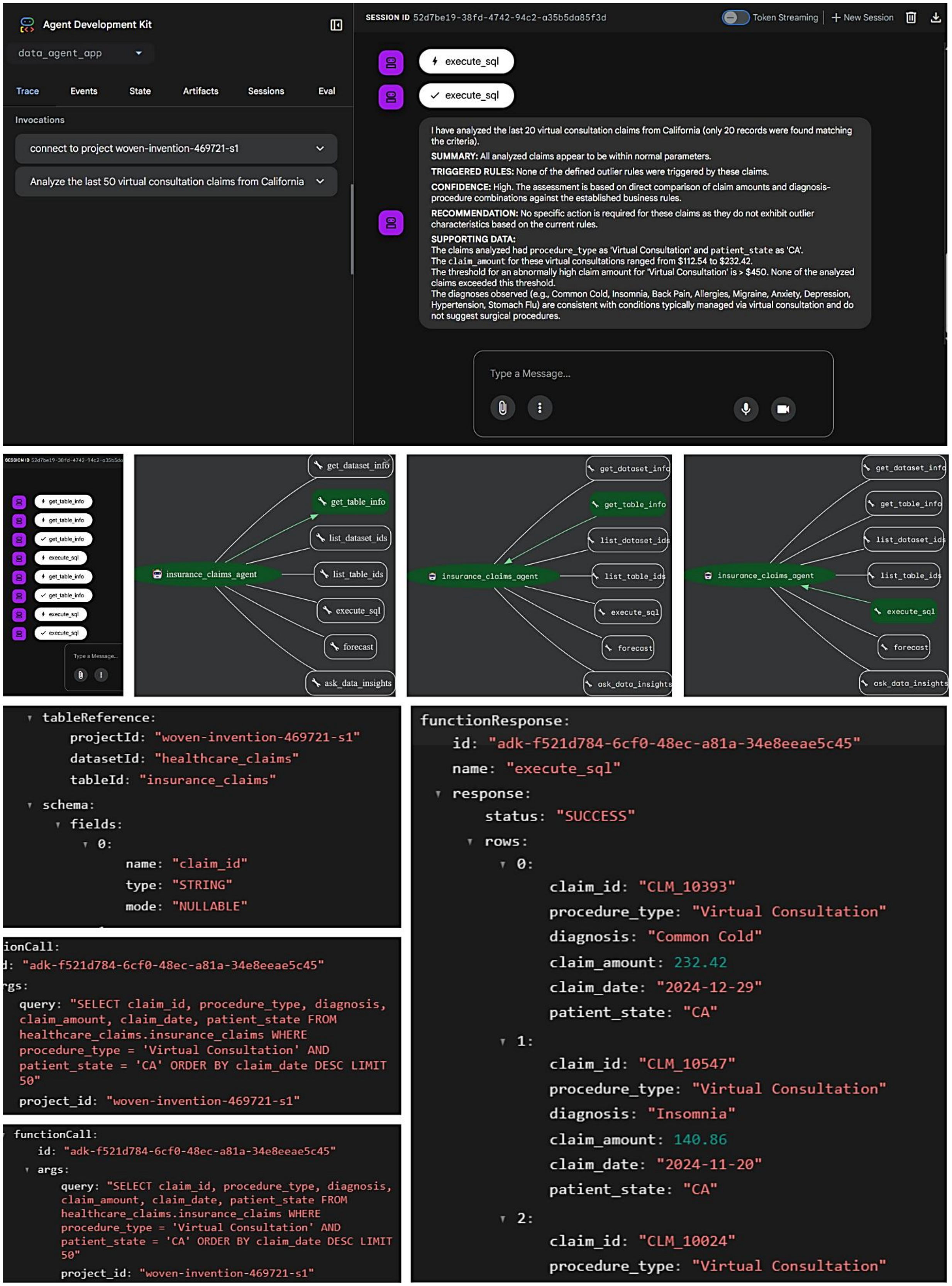

**Figure 6.** A complete view of how a single user query triggers the entire system architecture from user interface through agent processing to database execution and structured response generation.

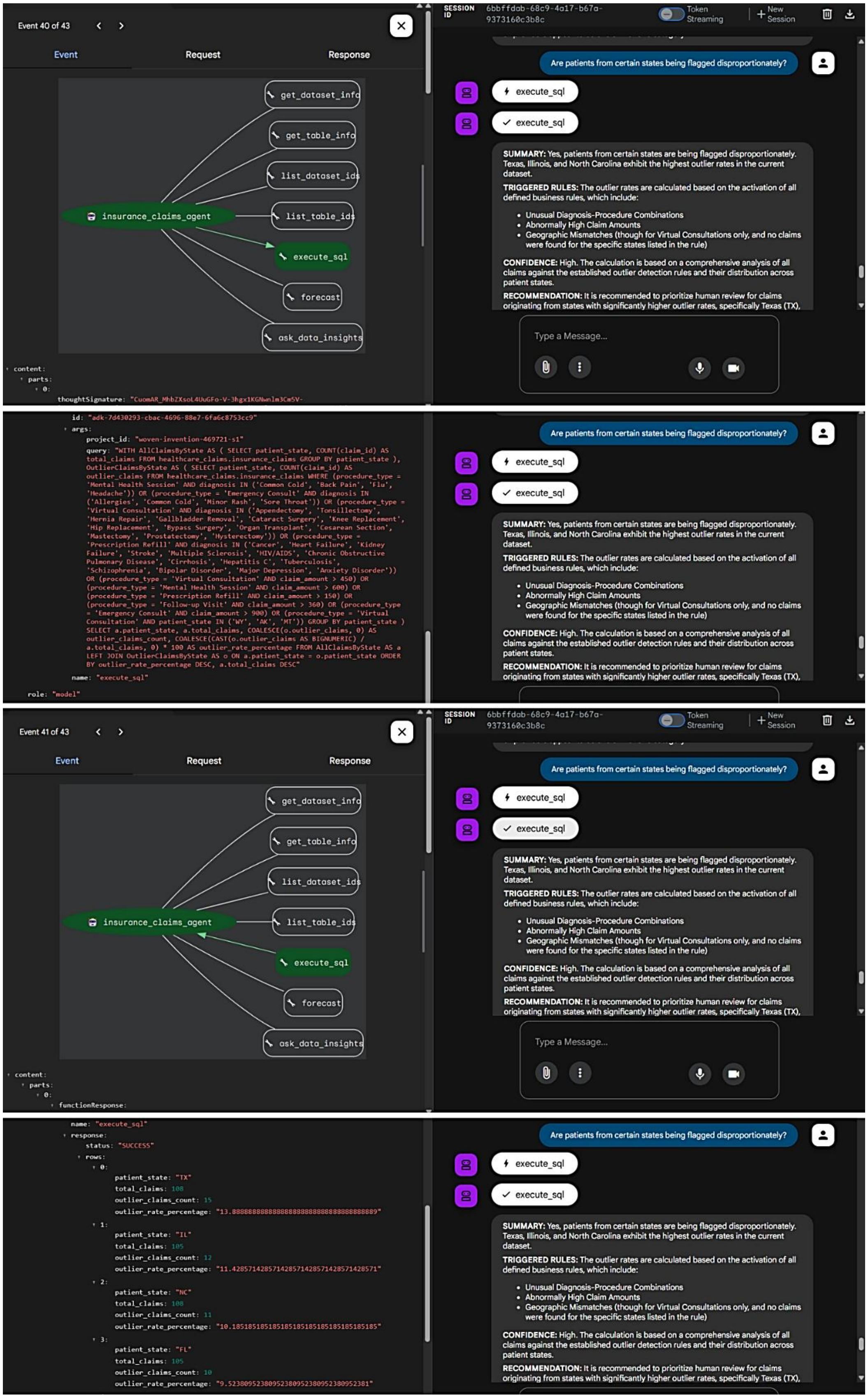

**Figure 7.** Execution flow: function calls → SQL query construction → statistical results.

Four panels in **Figure 7** illustrate the complete execution flow of a query, from backend processing to the final results. The first panel (Event 40) displays the function call graph, where the insurance_claims_agent initiates function call from a set of functions including get_dataset_info, get_table_info, list_dataset_id, and list_table_id. This ultimately leads to the execution of an SQL query that analyzes geographic patterns in claim outliers. On the right side, the panel shows the agent's summary response. The second panel reveals the actual SQL query being constructed. It demonstrates complex logic that analyzes claims by patient state, counting total and outlier claims while applying business rules for unusual diagnosis-procedure combinations, abnormally high claim amounts, and geographic mismatches. The query also includes specific filters for various procedure types and their associated diagnoses. The third panel (Event 41) shows the continuation of the function execution, specifically the execute_sql function call. This panel highlights the systematic approach the agent takes to gather dataset information before constructing and executing queries. The fourth panel presents the raw JSON response data with specific numeric results: Texas has 108 total claims with 15 outliers (13.89% rate), Illinois has 105 claims with 12 outliers (11.43% rate), North Carolina has 108 claims with 11 outliers (10.19% rate), and Florida has 105 claims with 10 outliers (9.52% rate). This data provides concrete support for the agent's analysis of geographic bias in claims flagging.

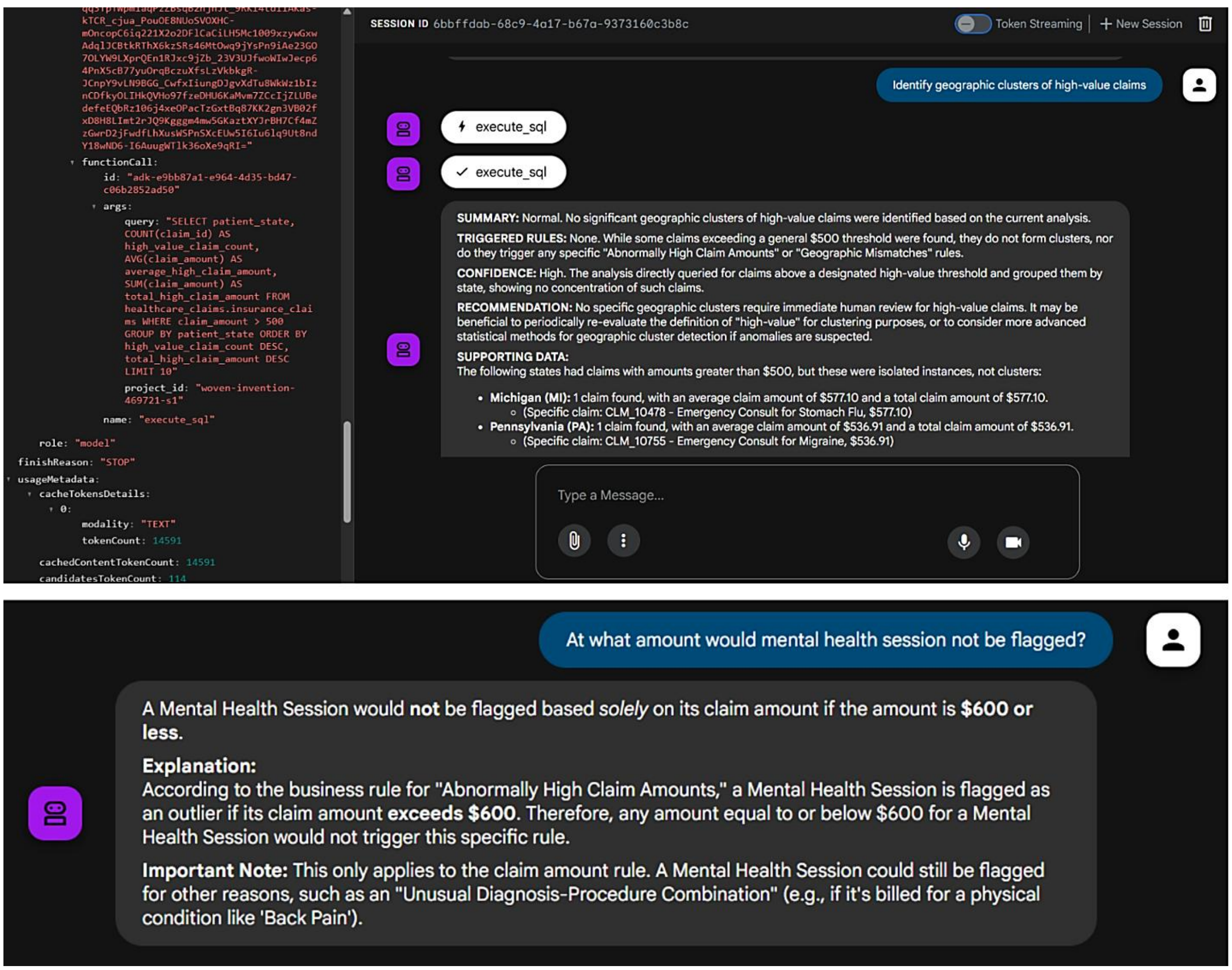

**Figure 8.** Agent's ability to provide precise, rule-based explanations with supporting evidence.

**Figure 8** demonstrates the system's capabilities in business rule transparency, multi-rule awareness, and data-driven responses through concrete examples. The top session displays complex SQL queries in the left panel, filtering for elevated claim amounts, performing statistical aggregations such as averages, counts, and percentages by region, and implementing outlier detection logic based on business rules. The right panel shows the AI's response to the user query, "Identify geographic clusters of high-value claims", where the agent concludes that no significant geographic clusters were identified after analyzing claim distribution patterns using the "Abnormally High Claim Amounts" rule. Supporting data includes examples such as a Michigan claim and a Pennsylvania claim. The bottom session addresses the Claims Threshold with the user query, "At what amount would mental health sessions not be flagged?" The AI responds that claims would not be flagged if the amount is $600 or below, based on the "Abnormally High Claim Amounts" rule.

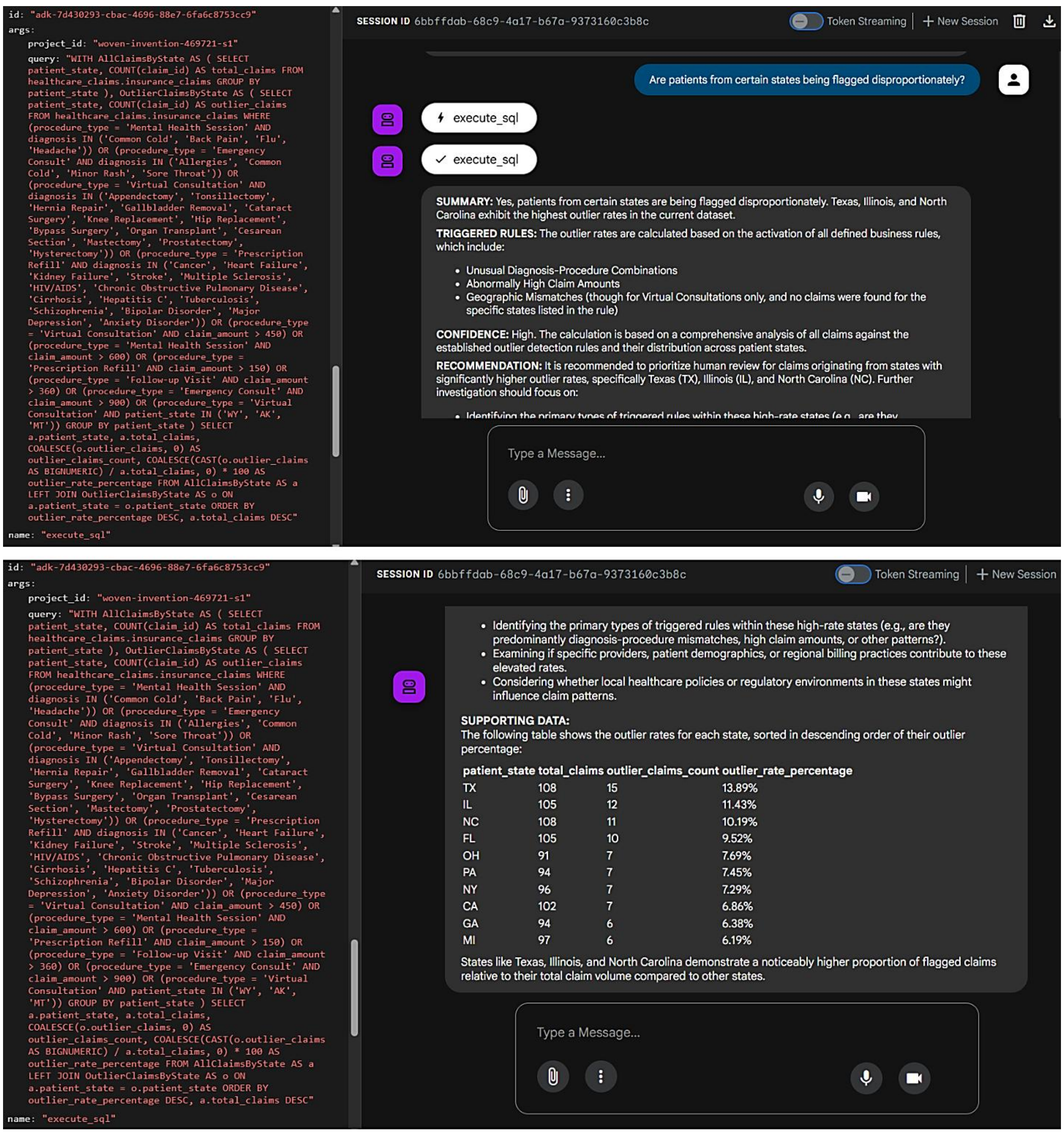

**Figure 9.** Figure 9 illustrates how the AI agent performs a series of backend steps, starting with SQL query generation, followed by state-based filtering, data aggregation, and sorting operations.

**Figure 9** illustrates how the AI agent translates business questions into database queries, applies domain-specific rules, and returns actionable insights about potential geographic bias in the claims flagging system. It presents two conversation sessions with the claims analysis agent. The left panel displays SQL queries being executed against the database, including filtering logic (e.g., WHERE clauses), aggregation functions (e.g., COUNT, AVG), and state-based filtering and grouping. The right panel shows user-agent interactions: in Session 1 (top), the user asks, "Are patients from certain states being flagged disproportionately?" and the AI responds by referencing specific outlier detection rules, identifying that geographic restrictions are contributing to disproportionate flagging in certain states. In Session 2 (bottom), the user inquires about outlier rates by state, and the AI provides a data table displaying outlier statistics broken down by state.

```
{
    "summary_metrics": {
        "row_count": 10,
        "outlier_detection_metric/mean": 3.4,
        "outlier_detection_metric/std": 2.065591117977289,
        "factual_accuracy_metric/mean": 2.6,
        "factual_accuracy_metric/std": 2.0655911179772892,
        "completeness_metric/mean": 3.4,
        "completeness_metric/std": 2.065591117977289,
        "trajectory_single_tool_use/mean": 0.1111111111111111,
        "trajectory_single_tool_use/std": 0.3333333333333333,
        "latency_in_seconds/mean": 14.726886909300083,
        "latency_in_seconds/std": 13.344001799725772,
        "failure/mean": 0.0,
        "failure/std": 0.0
    },
```

**Figure 10.** The evaluation reveals an AI agent with inconsistent performance, while it doesn't crash, it struggles with accuracy and proper tool usage. The high standard deviations across all custom metrics suggest the agent performs very differently across various types of insurance claims scenarios

**Figure 10** presents a JSON summary of evaluation metrics from the AI agent's performance test on insurance claims analysis tasks. The evaluation dataset contained 10 test cases. The Outlier Detection Metric evaluates how well the AI identifies unusual insurance claims and explains its reasoning, with a mean score of 3.4 out of 5 and a standard deviation of 2.07. The moderate mean score suggests mixed performance, while the high variability indicates that some cases were handled excellently and others poorly. The Factual Accuracy Metric measures whether the AI's responses contain correct factual information compared to the ground truth, with a mean of 2.6 out of 5 and a standard deviation of 2.07. This below-average score points to significant accuracy issues across test cases. The Completeness Metric assesses whether the AI provides all requested information, with a mean of 3.4 out of 5 and a standard deviation of 2.07, indicating moderate performance and occasional omission of important details. The Trajectory Single Tool Use metric measures the proper usage of the "list_table_ids" tool, showing a very low mean score of 0.11 out of 1 and a standard deviation of 0.33, meaning the agent rarely used this tool correctly when needed. Latency metrics show a mean response time of approximately 14.7 seconds with a standard deviation of about 13.3 seconds, reflecting high variability in response durations. Finally, the Reliability metric reports a failure rate of 0.0%, indicating no system failures occurred during testing.

**Figure 11** presents a detailed breakdown of a single evaluation case from the insurance claims evaluation system. The test case structure begins with the prompt and the question posed to the AI agent being evaluated. The reference, or ground truth, lists three expected outlier reasons, which serve as the benchmark for evaluation. In analyzing the AI agent's response, it correctly identified one outlier reason. The response time was approximately 11.78 seconds, with no failures. The tool usage trajectory reveals the AI's interactions with tools: the get_table_info tool accessed the "claims_data" table from the "insurance_claims" dataset under project ID "woven-invention-469721-s1", and the execute_sql tool ran an SQL query to count outlier reasons by selecting and grouping the relevant data accordingly. Detailed metric evaluations show that the Outlier Detection Metric scored 1.0 (poor) because the AI identified only one correct outlier reason, missed two others, and added a vague, non-reference category. The Factual Accuracy Metric also scored 1.0 (poor) due to fabricated numerical data not provided in the reference. The Completeness Metric scored 1.0 (poor) because the response was critically incomplete, listing only one of the three required outlier reasons, omitting two, and including a non-reference item instead of completing the list. Lastly, the Trajectory Single Tool Use metric scored 0.0, indicating that although the agent used tools, it did not utilize the specific list_table_ids tool that the evaluation was designed to assess.

```
"pointwise_metrics": [
    {
        "prompt": "What are the most common outlier reasons in the insurance claims dataset?",
        "reference": "The most common outlier reasons should include unusual diagnosis-procedure combinations, abnormally high claim amounts, and geographic
        coverage restrictions.",
        "response": "The most common outlier reasons in the insurance claims dataset are:\n\n1.  **Unusual diagnosis-procedure combination**: 38 occurrences\n2.
        **Suspicious claim pattern**: 14 occurrences",
        "latency_in_seconds": 11.78403822600012,
        "failure": 0,
        "predicted_trajectory": [
            {
                "tool_name": "get_table_info",
                "tool_input": {
                    "table_id": "claims_data",
                    "dataset_id": "insurance_claims",
                    "project_id": "woven-invention-469721-s1"
                }
            },
            {
                "tool_name": "execute_sql",
                "tool_input": {
                    "project_id": "woven-invention-469721-s1",
                    "query": "SELECT outlier_reason, COUNT(*) as reason_count FROM `woven-invention-469721-s1.insurance_claims.claims_data` WHERE is_outlier =
                    TRUE GROUP BY outlier_reason ORDER BY reason_count DESC LIMIT 5;"
                }
            }
        ],
       "outlier_detection_metric/explanation": "The AI correctly identified one outlier reason, but missed two other common and specific reasons from the
       reference (abnormally high claim amounts, geographic coverage restrictions). It replaced them with a vague 'Suspicious claim pattern' and added
       unrequested occurrence counts, demonstrating poor detection accuracy and limited mechanistic interpretability for its second point.",
       "outlier_detection_metric/score": 1.0,
       "factual_accuracy_metric/explanation": "The AI response omits two of the three common outlier reasons mentioned in the reference ('abnormally high claim
       amounts' and 'geographic coverage restrictions') and adds one not present ('suspicious claim pattern'). Furthermore, the AI fabricates numerical
       occurrence data (e.g., '38 occurrences', '14 occurrences') that is not provided in the reference at all. These are significant factual errors and
       fabrications.",
       "factual_accuracy_metric/score": 1.0,
       "completeness_metric/explanation": "The AI's answer is critically incomplete as it only lists one of the three common outlier reasons provided in the
       reference ('unusual diagnosis-procedure combination') and completely omits 'abnormally high claim amounts' and 'geographic coverage restrictions'. It
       also adds an item ('suspicious claim pattern') not present in the reference.",
       "completeness_metric/score": 1.0,
       "trajectory_single_tool_use/score": 0.0
    },
    {
```

**Figure 11.** This figure illustrates a systematic way to test and measure how well an AI agent answers questions by using structured evaluation tools. The system tracks what the agent actually did, which tools it used, what database it queried, how long it took, etc. This whole setup is used to quantitatively measure the performance of the AI agent, how accurately it retrieves information and responds to user questions. In the example shown, the agent is asked to identify common outlier reasons in the insurance claims dataset. Its response, generated through database queries and tool calls, is evaluated against a reference answer. The system tracks execution details, including invoked functions and query results. According to automated evaluation, AI correctly identified one reason while missing two others and fabricating numerical occurrences do not present in the reference. This framework allows for systematic measurement of how well the agent retrieves correct information and presents it in response to natural language queries.

## 4. DISCUSSION

The implementation and evaluation of the AI agent for insurance claims analysis demonstrate an integration of agentic systems with enterprise data infrastructure. This work provides a proof-of-concept to establish a robust, evaluable framework for deploying LLM-powered agents in data-rich domains. The discussion that follows synthesizes the key outcomes, highlighting how the system's architecture addresses core challenges in data accessibility, quality assurance, and transparent decision-making.

### 4.1. Synthesizing Natural Language Interfaces with Enterprise Data Warehouses

A primary achievement of this work is the integration of a natural language interface with a structured BigQuery data warehouse. As evidenced in the user interactions, the system bridges the semantic gap between user intent and complex SQL query execution. This capability fundamentally democratizes data access, enabling non-technical users to perform sophisticated data retrieval and analysis without requiring specialized knowledge of the underlying schema or query language.

The system's effectiveness hinges on the Agent Development Kit (ADK) Framework and its Safe Query Builder, which provide a controlled environment for query generation. This design prevents malformed queries and enforces data security through parameterized, role-limited access. The result is a conversational interface that is not only user-friendly but also reliable and secure, making complex data warehouses interactively queryable.

### 4.2. The Critical Role of an Automated Evaluation Framework

A cornerstone of this system's design is its evaluation framework. The ability to automatically benchmark agent performance against a "golden" test dataset introduces a rigor to agent development and maintenance. This framework moves quality assurance from subjective assessment to quantitative measurement. It enables developers to:

- Benchmark Performance: Objectively compare different agent versions or LLM models.
- Detect Regressions: Identify performance degradation during updates before deployment.
- Drive Iterative Improvement: Pinpoint specific weaknesses to focus development efforts.

This continuous feedback loop is essential for building trust in autonomous systems and ensuring their performance meets production standards over time.

### 4.3. Achieving Transparent and Auditable Reasoning through Interpretable Design

In high-stakes domains like insurance and healthcare, the "black box" nature of AI is a significant barrier to adoption. This system directly addresses this challenge through its Multi-Layered Interpretable Reasoning architecture. The agent's decision-making process is not opaque; it is systematically decomposed into Input Feature Analysis, Rule Activation, Decision Pathway, and Confidence Scoring. The analysis of claim CLM_10668 is a paradigmatic example. The agent did not simply state the claim was an outlier; it provided an explanation, citing the specific business rule violated ("Unusual Diagnosis-Procedure Combinations"), supported by quantitative evidence (the diagnosis appeared in only 9.95% of similar procedures). This 4-Layer Interpretability Framework provides a clear audit trail, which is crucial for regulatory compliance, handling appeals and building user confidence. It ensures that every data-driven decision can be traced and justified.

### 4.4. Statistical and Contextual Analysis

While many systems rely solely on static business rules, this agent enhances its analytical depth by integrating a Statistical Comparator. This component allows the system to “quantify deviation from normative behavior” by comparing individual claims against aggregated historical data. This is evident in the agent's ability to calculate that a claim amount was 67.47% below the average, providing a more nuanced context than a simple threshold check.

Furthermore, the system demonstrates multi-rule awareness, as seen when analyzing geographic bias. It can execute complex queries that consider the interplay of different rules (e.g., geographic restrictions combined with high-cost flags) to answer holistic business questions like “Are patients from certain states being flagged disproportionately”? This ability to perform normative analysis and synthesize across multiple rules represents a significant advancement over basic, siloed rule-checking systems.

### 4.5. Future works

Future work will focus on integrating machine learning models for anomaly detection to complement the rule-based system and identify emerging, sophisticated fraud schemes.